\documentclass[conference]{IEEEtran}
\IEEEoverridecommandlockouts
\usepackage{stfloats}
\usepackage{cite}
\usepackage{amsmath,amssymb,amsfonts}
\usepackage{algorithmic}
\usepackage{graphicx}
\usepackage{textcomp}
\usepackage{xcolor}
\usepackage{hyperref}
\usepackage{multirow}
\usepackage{threeparttable}
\usepackage{booktabs}
\usepackage{makecell}
\usepackage{colortbl}
\usepackage{tabularx}
\usepackage{bbding}
\usepackage{tabstackengine}
\usepackage{lipsum}
\def\BibTeX{{\rm B\kern-.05em{\sc i\kern-.025em b}\kern-.08em
    T\kern-.1667em\lower.7ex\hbox{E}\kern-.125emX}}

\newcommand{\CircuitType}{3}
\newcommand{\OELow}{2.32}
\newcommand{\OEHigh}{26.6}

\begin{document}

\title{A Large Language Model-based Multi-Agent Framework for Analog Circuits' Sizing Relationships Extraction}
% \author{\IEEEauthorblockN{Anonymous Authors}}
\author{\IEEEauthorblockN{Chengjie Liu}
\IEEEauthorblockA{\textit{School of Electronic Science}\\ \textit{and Engineering, Nanjing University} \\
\textit{National Center of Technology}\\ \textit{Innovation for EDA}\\
Nanjing, China \\
cjliu\_phd@smail.nju.edu.cn}
\and
\IEEEauthorblockN{Weiyu Chen}
\IEEEauthorblockA{\textit{School of Electronic Science}\\ \textit{and Engineering, Nanjing University} \\
\textit{National Center of Technology}\\ \textit{Innovation for EDA}\\
Nanjing, China \\
wychen@smail.nju.edu.cn}
\and
\IEEEauthorblockN{Huiyao Xu}
\IEEEauthorblockA{\textit{Nanjing Tech University} \\
\textit{National Center of Technology}\\ \textit{Innovation for EDA}\\
Nanjing, China \\
202121018094@njtech.edu.cn}
\and
\IEEEauthorblockN{Yuan Du}
\IEEEauthorblockA{\textit{School of Electronic Science}\\ \textit{and Engineering, Nanjing University} \\
Nanjing, China \\
yuandu@nju.edu.cn}
\and
\IEEEauthorblockN{Jun Yang}
\IEEEauthorblockA{\textit{School of Integrated Circuits}\\ \textit{Southeast University} \\
\textit{National Center of Technology}\\ \textit{Innovation for EDA}\\
Nanjing, China \\
dragon@seu.edu.cn}
\and
\IEEEauthorblockN{Li Du}
\thanks{*Li Du is the corresponding author.}
\IEEEauthorblockA{\textit{School of Electronic Science}\\ \textit{and Engineering, Nanjing University} \\
Nanjing, China \\
ldu@nju.edu.cn}
}
\maketitle

\begin{abstract}
In the design process of the analog circuit pre-layout phase, device sizing is an important step in determining whether an analog circuit can meet the required performance metrics. 
Many existing techniques extract the circuit sizing task as a mathematical optimization problem to solve and continuously improve the optimization efficiency from a mathematical perspective.
But they ignore the automatic introduction of prior knowledge, fail to achieve effective pruning of the search space, which thereby leads to a considerable compression margin remaining in the search space. 
To alleviate this problem, we propose a large language model (LLM)-based multi-agent framework for analog circuits' sizing relationships extraction from academic papers. The search space in the sizing process can be effectively pruned based on the sizing relationship extracted by this framework. Eventually, we conducted tests on {\CircuitType} types of circuits, and the optimization efficiency was improved by $\OELow \sim \OEHigh \times$. This work demonstrates that the LLM can effectively prune the search space for analog circuit sizing, providing a new solution for the combination of LLMs and conventional analog circuit design automation methods. 
\end{abstract}


\section{Introduction}
Analog circuits are an essential component of integrated circuits (ICs). Its design process consists of the pre-layout and layout phases. 
The sizing process in its pre-layout phase requires manual repetitive parameter tuning to achieve higher circuit performance as much as possible, resulting in a considerable consumption of repetitive manpower. 
To address this issue, many works have implemented automatic parameter generation or optimization through deep learning\cite{dongcktgnn}, reinforcement learning\cite{wang2020gcn}, or optimization algorithms\cite{pmlr-v80-lyu18a,zhang2021efficient,9580454}. 
% Among them, for optimization algorithms, how to enable it to optimize analog circuits with more devices with higher search efficiency and success rate is a key research question.

Among them, the optimization algorithm transforms the sizing problem of analog circuits into a black-box optimization problem.
The goal of optimization algorithms is to achieve higher optimization efficiency and accuracy rate for complex circuits with more devices from a mathematical perspective. However, analog circuits are not completely black-box optimization problems. Introducing knowledge related to analog circuits into the optimization process to provide optimization directions or pruning the search space can be effectively combined with existing optimization algorithms, thereby further improving the efficiency of the optimization algorithm from the perspective of analog circuit design. 

Recently, many efforts\cite{liu2024ladac,lai2024analogcoder,yin2024ado,liu2024ampagentllmbasedmultiagentmultistage} have explored how to utilize large language models (LLMs) to enhance the efficiency of analog circuit design.
And they have been fully verified that LLMs can improve the efficiency and accuracy rate\cite{yin2024ado,liu2024ampagentllmbasedmultiagentmultistage} in the analog circuit sizing process by providing initial values for conventional optimization algorithms. 
However, the existing general LLMs, such as GPT-4o\footnote{\url{https://chatgpt.com/}}, DeepSeek R1\footnote{\url{https://www.deepseek.com/}}, and Claude3\footnote{\url{https://claude.ai}}, have not undergone specialized training for analog circuit design\cite{liu2024ampagentllmbasedmultiagentmultistage}. This results in their inability to understand complex or relatively new-structured analog circuits, and they tend to either be only applicable to classic analog circuits\cite{liu2024ladac,yin2024ado} or require a large number of fixed computational processes\cite{liu2024ampagentllmbasedmultiagentmultistage} to provide relatively good initial values. 
Thus, directly using general LLMs to provide initial values in the analog circuit sizing process to accelerate the optimization algorithm efficiency is limited. 

% Therefore, to further utilize the reasoning ability of the general LLMs as prior knowledge to improve efficiency in the sizing of a wider variety of analog circuits, we have explored a method to enhance the optimization algorithm efficiency from the perspective of search space pruning by leveraging LLMs.
In order to continuously and fully leverage the reasoning capabilities of the general LLMs as prior knowledge and be applicable to a broader range of circuits, we thus commence from the perspective of search space pruning to further enhance the efficiency of the optimization algorithm.

In this paper, we will introduce a large language model-based multi-agent framework for analog circuits' sizing relationship extraction. This method converts the circuit description from the paper into sizing relationships, thereby pruning the search space and improving the search efficiency. Unlike the methods that require a large amount of computation\cite{liu2024ampagentllmbasedmultiagentmultistage} or have a strong analog-circuit-domain zero-shot answering ability\cite{yin2024ado} for providing initial values, sizing relationships extraction relies more on the text reasoning ability that LLM is good at. It eliminates the reasoning process of mathematical formulas in the field of analog circuits and significantly improves the reliability of the LLMs' output.

This method is a universal method applicable to multiple common analog circuit modules, such as Operational Amplifier (OPA), Bandgap Reference (BGR), and Low DropOut regulator (LDO). We conducted tests on these {\CircuitType} types of circuits, and the optimization efficiency was improved by $\OELow \sim \OEHigh \times$ and successfully optimized the circuits that conventional optimization algorithms failed in limited iterations.
Overall, this method provides an innovative solution for improving the automation efficiency of analog circuit sizing.
It also provides a solution for high-quality analog circuit data generation in the current background of analog circuit data shortage.  

Our paper is organized as follows: Section II will introduce the background and the reasons why we adopt general LLMs instead of training a Domain-specific LLM. Section III will introduce the specific workflow. Section IV will specifically describe the experimental results. Finally, we will summarize and discuss our work.

\begin{figure*}[htbp]
\centering 
\includegraphics[width=6.4in]{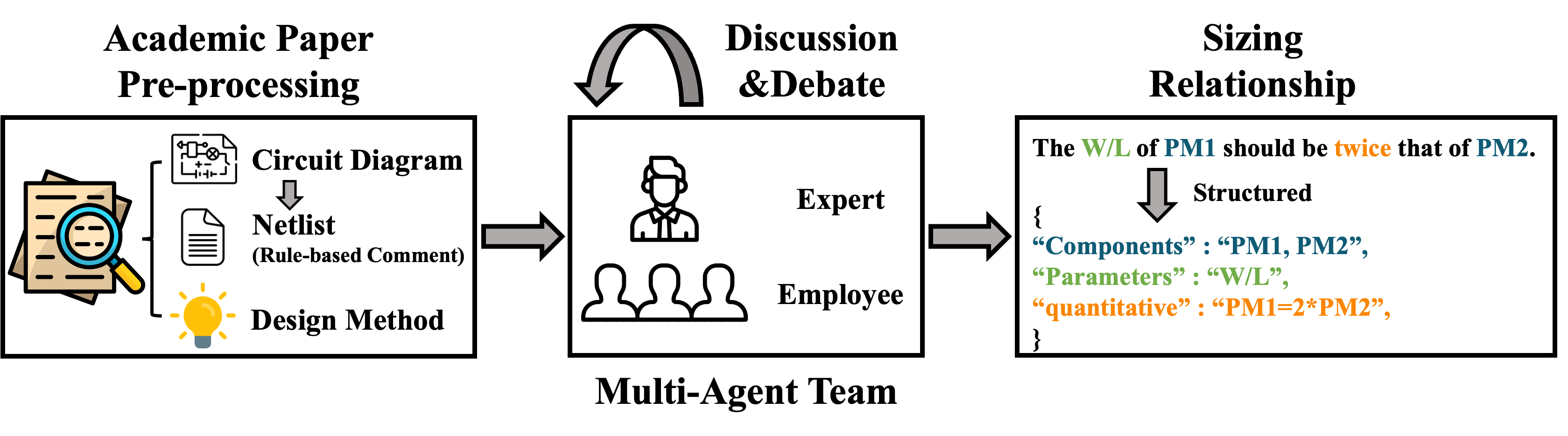}
\caption{The overview of the sizing relationship extraction} 
\label{Fig:overview}
\end{figure*}
\section{Preliminaries}
\subsection{LLMs and related research}
Large Language Models (LLMs)  are a type of artificial intelligence technology based on deep learning neural networks that contain billions (even more) of parameters\cite{zhao2023survey}, which are trained on massive text data\cite{shanahan2024talking}. These allow them to grasp the nuances, context, and complexity of human language. As a result, LLMs can perform a wide range of language-related tasks, such as translation, summarization, question answering, and content generation, with a remarkable degree of proficiency. 

Many studies\cite{xi2025rise} have carried out the use of LLM as an agent to make decisions on various tasks, and it has accomplished tasks in fields such as autonomous driving\cite{wen2023dilu}, video games\cite{wang2023voyager}, and even Nobel-Prize-level chemical experiments\cite{boiko2023autonomous} outstandingly.  

To further enable LLMs to handle more complex system-level tasks, many studies\cite{guo2024large} have investigated the adoption of multi-agent solutions. Through cooperation and debate among multiple agents, the collective wisdom of multiple agents has been fully brought into play. Through this type of method, multi-agent systems have accomplished engineering problems that require the expertise of many more human experts, such as software development\cite{qian2023communicative}, the crystallinity of MOFs
and COFs optimization\cite{zheng2023chatgpt}, and medical diagnosis assistance\cite{tang2023medagents}.

Through the breakthroughs of the above LLM agents in various fields, we are confident in using a set of LLM-based multi-agent systems to handle engineering problems in analog circuit design.

\subsection{LLMs for Integrated Circuits}
In the field of integrated circuits, LLMs have been widely applied in digital circuit design, such as RTL verification \cite{hu2024uvllm}, bug locations\cite{yao2024location}, and even CPU generation\cite{wang2024chatcpu}, significantly improving the development efficiency of digital circuits. Compared with LLM for digital circuits, LLM for analog circuits has the problem of a shortage of training data. Many existing training-based LLMs for analog circuit auto-design techniques\cite{chang2024lamagic,chen2024artisan} rely on self-organized data. However, compared with other datasets used for the training of LLMs \cite{liu2024datasets}, where a single pre-trained dataset can reach hundreds of terabytes in size, and even a single dataset in the supervised fine-tuning stage can contain millions of data pairs, the data collected in \cite{chang2024lamagic, chen2024artisan} is of a very small quantity. This limited data leads to the LLMs they trained being capable of solving only circuits with fixed structures or potentially encountering the problem of overfitting.

Therefore, how to combine LLMs with the existing analog circuit automation design methods and generate high-quality diverse analog circuit data is what makes more significant contributions of LLMs for analog circuits at present. 
In this work, we focus on accelerating the process of extracting actual circuit data from papers based on existing general LLMs: through the extraction of the sizing relationship of the circuits by LLMs, pruning of the search space is carried out, thereby improving the efficiency and reliability of the sizing process, and ultimately obtaining high-quality analog circuit data. 

\section{Overview}
The overview of the multi-agent framework is shown in Fig.\ref{Fig:overview}. 
The papers will first undergo preprocessing to extract the circuit diagram, netlist, and design method related to the circuits mentioned in the papers. These data will be taken together as input to the multi-agent team for iterative discussion and analysis. 
After reaching a fixed number of iterations, the conclusion information summarized by the team will be sent as the extracted sizing relationship to the conventional sizing algorithm for optimizing the circuit netlist. 
Next, a detailed elaboration will be provided for each module.
\subsection{Paper Pre-Processing}
Since the input of existing LLMs is generally text or text along with images, in order to extract the sizing relationship information from the papers, we need to preprocess the PDF format files of the papers into text first. Here, we have adopted the commercial tool Mathpix\footnote{\url{https://mathpix.com}}. After this step, we obtained the circuit diagrams in the papers and the design information related to the circuits.  

After obtaining the circuit diagrams, in order to convert them into simulatable netlists, we adopted a similar technique as in \cite{tao2024amsnet}. We will identify the devices, connection relationships, and text in the pictures through CV solutions such as Yolo-V8\cite{yolov8} and OCR, thereby restoring the corresponding netlist from the circuit diagrams in the papers. 
Next, a script specifically designed for module detection will identify and detect special modules in the netlist and recognize the common and necessary ones, such as differential pairs and current mirrors. Through this method, the pressure on multi-agents for fine-grained module recognition can be alleviated, and the accuracy of the final sizing relationship extraction can be improved.  
\subsection{Multi-Agent Team}
% 介绍Multi-Agent的优势简单过一下，讲Multi-Agent结构，接下来进一步讲讨论过程，给个例子图
As mentioned in Part A of Section II, the multi-agent system has achieved good results in multiple engineering tasks. Here, we also adopt a multi-agent system to achieve the accurate extraction of the sizing relationship of the analog circuits in the papers. 

The construction methods of the multi-agent system mainly unfold around three aspects\cite{guo2024large}, including 1) \textit{Communication Paradigms}: the styles and methods of interaction between agents; 2) \textit{Communication Structure}: the organization and architecture of communication networks within the multi-agent system; and 3) \textit{Communication Content}: what kind of message is exchanged between agents.

\textit{Communication Paradigms} we adopt include \textbf{cooperation} and \textbf{debate}: First of all, these models need to cooperate to discuss the correct sizing relationship. Compared with the single-agent system, the multi-agent system is more likely to achieve comprehensive coverage of the extraction of the sizing relationship. Secondly, since existing general models have not received specialized training for analog circuits, a single agent may have the possibility of extracting the relationship wrongly, which leads to the occurrence of opinion conflicts among multiple agents. At this time, a debate is required to allow agents with different opinions to discuss, making them pay more attention to extracting strong evidence based on the papers, thereby making the extracted sizing relationship more reliable. 

\textit{Communication Structure} we adopt is the \textbf{Shared Message Pool}, which is proposed in \cite{hong2023metagpt}. These agents can see the sizing relationships and specific ideas extracted by each other, improving the communication efficiency among agents.

\textit{Communication Content} includes the \textbf{structured extraction results} of sizing relationships and the \textbf{thinking processes} involved in extracting these relationships. By sharing the thinking processes, agents can better analyze the reasons for extracting sizing relationships from each other, thereby further enhancing the reliability of the overall agent system in extracting sizing relationships. 

In addition to the common multi-agent construction mentioned above, we assigned different models to agents of different roles. In this work, We divided the roles into one expert and multiple employees. In terms of personnel allocation, we found that if the agents were all focused on the discussion, they only tended to extract the sizing relationship from the paper in the first round (before the discussion), and in the remaining iteration rounds, they only focused on the discussion, resulting in the possible problem of missed extraction of the sizing relationship. Therefore, we ultimately chose to have the Expert Agent not participate in the discussion and propose new sizing relationships based on the academic paper in each iteration round, thereby making the coverage of the extracted sizing relationships more extensive.  Combined with the characteristics of each role, the expert agent is more inclined to use an \textbf{LLM with a long thought chain}, while the employee is inclined to use a \textbf{relatively low-cost LLM}. The specific LLM selection will be elaborated in detail in the experimental section. 

Through the above configuration, this multi-agent team can stably and efficiently extract the sizing relationships of various circuits in the papers to be reproduced accurately. A detailed extraction process will be described in the experimental section. 

\subsection{Sizing Relationship Format}
The sizing relationship we extracted mainly includes three key pieces of information: device name, related parameters, and quantitative relationship, as shown in Fig.\ref{Fig:overview}. The device name will be used to match the corresponding device in the netlist for matching. The related parameters will be used to specify what parameters have the relationship set. The quantitative relationship specifically records the specific quantitative magnitude between different device parameters. 
The output of the multi-agent team will be structured and output in the above format, thus cascading with the subsequent optimization algorithm.

\section{Experiment}

In this section, we will present in detail an example of extracting sizing relationships using this method. Furthermore, we extracted the sizing relationships for three circuits and compared them with the efficiency and accuracy of the optimization algorithm when no sizing relationships were provided. 
\begin{table}[t]
    \renewcommand{\arraystretch}{1.2}
    \centering
    \caption{The configurations of the multi-agent team}
    \label{tab:configuration}
    \resizebox{\linewidth}{!}{ 
        \begin{tabular}{c|c c c|c c c}
        \toprule
        \multirow{2}{*}{} & \multicolumn{3}{c|}{\textbf{Expert Agent}} & \multicolumn{3}{c}{\textbf{Employee Agent}} \\
        % \cline{2-7}
         & Model & Num. & Temp. & Model & Num. & Temp. \\
        % \hline
        \midrule
        CFG 1 & GPT-o1 & 1 & $\times$ & GPT-4o& 3 & 0.5\\
        % \hline
        CFG 2 & GPT-o1 & 1 & $\times$ & Claude3.5-opus& 3 & 0.5\\
        \bottomrule
        \multicolumn{7}{l}{$\times$GPT-o1 does not support temperature adjustment.  }
        \end{tabular}
    }
\end{table}
\subsection{Experiment Setting}
In this section, we will provide a detailed explanation of the experimental setup. Initially, it is important to note that the proposed method is designed for general purposes, and the application of LLMs from various sources, provided they are of comparable levels, will not significantly alter the outcomes of the multi-agent team. To substantiate this consideration, we employed two distinct state-of-the-art general LLMs configurations for the multi-agent team, as detailed in Table.\ref{tab:configuration}. \footnote{Because among the current long reasoning LLMs, only GPT-o1 has the ability of image understanding, GPT-o1 was selected as the only expert agent here.} The number of discussion rounds of the multi-agent team is set to 5. The multi-agent team is based on LangChain\footnote{\url{https://python.langchain.com/}}.

In terms of the optimization algorithm used to perform sizing, the running CPU is Intel$^\circledR$ Xeon$^\circledR$ CPU E5-2698 v4 @ 2.20GHz, and the development platform is based on Tsinghua Electronic Design (TED)\cite{TED}. The optimization algorithm adopts MACE\cite{pmlr-v80-lyu18a}, its maximum number of iteration rounds is set to 100, the batch number is set to 40, and its initial point is set to 40.
The academic papers we selected for sizing extraction are \cite{ueno2009300,ferreira2007ultra,leung2003capacitor}.
\begin{figure*}[!t]
    \centering
    \includegraphics[width=\linewidth]{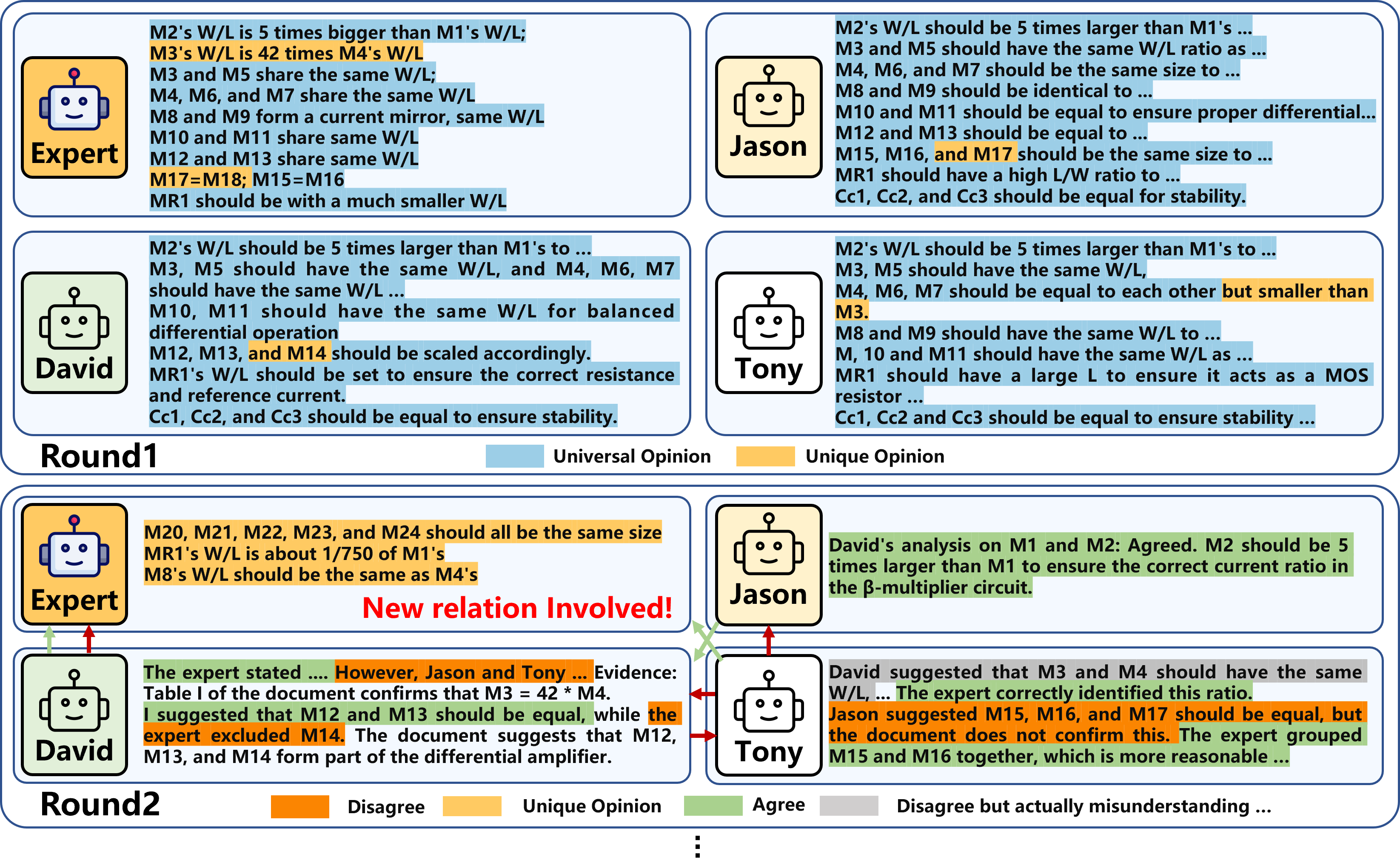}
    \caption{Examples when extracting sizing relationship from \cite{ueno2009300}}
    \label{fig: Discussion Example}
\end{figure*}
\subsection{Extraction Case}

% \begin{figure}
%     \centering
%     \includegraphics[width=\linewidth]{Experiment/circuit_diagram.png}
%     \caption{Caption}
%     \label{fig:CD}
% \end{figure}
To show the extraction process of the multi-agent team more explicitly, we have presented some of the contents output by the multi-agent team in the first two rounds when extracting sizing relationship from Paper \cite{ueno2009300} under CFG1, as shown in Fig.\ref{fig: Discussion Example}. 
% The circuit diagram in \cite{ueno2009300} after pre-processing is shown in Fig.\ref{fig:CD}.

Some of the thinking processes, as well as the opposing and agreeing viewpoints, have been omitted due to space limitations. However, it can be seen that in the first round, each agent extracted a considerable number of similar sizing relations, but there were some differences in each detail. It is worth noting that the Expert was the only one to propose the quantitative relationship of M3 and M4, which was calculated based on the parameter table in the paper. In the second round, each agent agreed and refuted each other's viewpoints. Furthermore, the Expert continued to introduce new sizing relationships and further paid attention to the relationship between the current mirror and other quantitative relationships.
\begin{table}[t]
\renewcommand{\arraystretch}{1}
\centering
\caption{Extraction Stability of the Multi-Agent Team}
\label{tab:sizing_relation_num}
\resizebox{\linewidth}{!}{ 
\begin{tabular}{c|c c|c c|c c}
  \toprule
  % \makecell{\#Valid Sizing\\Relationship}
  \multirow{2}{*}{}&\multicolumn{2}{c|}{\textbf{BGR}\cite{ueno2009300}} & \multicolumn{2}{c|}{\textbf{OP}\cite{ferreira2007ultra}} & \multicolumn{2}{c}{\textbf{LDO}\cite{leung2003capacitor}} \\ 
  % \cline{2-7}
  &CFG1 & CFG2 & CFG1 & CFG2 & CFG1 & CFG2 \\ 
  % \hline
  \midrule
  Attempt1& 11 & 12 &5&4& 10& 11\\ 
  % \hline
  Attempt2& 11 & 11&3&5& 9& 10\\ 
  % \hline
  Attempt3& 12 & 12&3&3& 10& 10\\ 
  % \hline
  Max Variation$\downarrow$& 1 & 1 & 2& 2& 1& 1\\ \bottomrule
\end{tabular}}
\end{table}
We also tested the extraction sizing relationship stability for these academic papers under two CFGs. We conducted three tests for each paper under each CFG and counted the number of final valid sizing relationships. Among them, a valid sizing relationship means that, for example, if some sizing extracted two items: \textit{M1 = M2, M2 = M3}, it is actually equivalent to \textit{M1 = M2 = M3}, and is actually counted as one valid piece sizing relationship. The specific experimental results are shown in Table.\ref{tab:sizing_relation_num}. It can be seen from Table.\ref{tab:sizing_relation_num} that regardless of which CFG is used, the number of extracted effective sizing relationships is relatively stable, proving the stability of the multi-agent team in this task and that the method proposed in this paper is a general one. As long as the LLM ability levels are comparable, it will not have an impact on the final result. 

\subsection{Efficiency Improvement}
\begin{table}[t]
    \renewcommand{\arraystretch}{1.2}
    \centering
    \caption{Circuit Optimization Objective Configuration\\ and Optimization Results Comparison}
    \resizebox{\linewidth}{!}{
    \begin{tabular}{c c c >{\columncolor{green!10}}c >{\columncolor{red!10}}c}
    % \hline
    \toprule
    &\textbf{Attribute} & \textbf{Target} & \textbf{\makecell{w/ sizing \\ Relationship}}&\textbf{\makecell{w/o sizing \\ Relationship}}\\ 
    % \midrule
    \hline
    \multirow{4}{*}{\makecell{\textbf{BGR}\cite{ueno2009300}\\@SMIC180nm}}&Vref& $0.5 \sim 1.3V$ & 1.24V & 1.39V\XSolidBrush\\ 
    % \cline{2-3}
    & TC$\downarrow$ & $<25$ppm& 23ppm& 28ppm\XSolidBrush\\ 
    % \cline{2-3}
    & PSR$\downarrow$ & $<-40$dB &-43dB& -45dB\\ 
    % \cline{2-3}
    & Power$\downarrow$ & $<1\mu$W & 242nW & 270nW\\ 
    \hline
    \multirow{8}{*}{\makecell{\textbf{OP}\cite{ferreira2007ultra}\dag\\@TSMC65nm}}&DC Gain$\uparrow$& $>70$dB & 71.1dB&55.7dB\XSolidBrush\\ 
    % \cline{2-3}
    & UGB$\uparrow$ & $>8$kHz&33kHz&332kHz\\ 
    % \cline{2-3}
    & Phase Margin$\uparrow$ & $>40^{\circ}$ &75$^{\circ}$&42.4$^{\circ}$\\ 
    % \cline{2-3}
    & PSRR$\uparrow$ & $>50$dB &63dB&53.0dB\\ 
    % \cline{2-3}
    & CMRR$\uparrow$ & $>60$dB &85.8dB&66.0dB\\ 
    % \cline{2-3}
    & Offset & $<5$mV &0.1mV&1mV\\ 
    % \cline{2-3}
    & Slew Rate$\uparrow$ & $>10$V/ms &27.5V/ms&381V/ms\\ 
    % \cline{2-3}
    & Power & $<1\mu$W &265nW&477nW\\ 
    % \midrule
    \hline
    \multirow{5}{*}{\makecell{\textbf{LDO}\cite{leung2003capacitor}\\@TSMC65nm}}&Load Regulation$\downarrow$& $<0.2$V/A &0.017V/A&0.046V/A\\ 
    % \cline{2-3}
    & Line Regulation$\downarrow$ & $<0.2$V/V&0.08V/V&1.1V/V\\ 
    % \cline{2-3}
    & Output Voltage & $1.05\sim1.15$V &1.09V&1.11V\\ 
    % \cline{2-3}
    & PSRR$\uparrow$ & $>50$dB &54.6dB&59.3dB\\ 
    % \cline{2-3}
    & Power$\downarrow$ & $<100\mu$W &46.5$\mu$W&84.1$\mu$W\\ 
    \bottomrule
    \multicolumn{5}{l}{\small $\dag$ Though performing well when no sizing relationship, oscillation occurs.}\\
    \end{tabular}
    }
    \label{tab:performance indicator}
\end{table}

In this part, we will compare the efficiency of the optimization algorithm with and without sizing relations.
The demand indicators of the three circuits and the optimization results with and without sizing relationships are presented in Table.\ref{tab:performance indicator}.  
% The optimization and constraint indicators of the three types of circuits are listed in Table.
It can be seen that after combining the sizing relationship, the optimization algorithm can achieve that all the indicators of the experimental circuit can be satisfied. However, using only the conventional algorithm, within fixed iterations, it is impossible to optimize the circuit parameters that meet all the indicator requirements. 
Additionally, it is worth mentioning that although the conventional optimization algorithm yields good final indicators when designing the OP, oscillation problems occur, which leads to the optimized circuit having no application value.

Moreover, the above experiments were carried out on two different process nodes, SMIC 180nm and TSMC 65nm, verifying the robustness of the method in terms of process migration. 
% The optimization algorithm adopts MACE\cite{pmlr-v80-lyu18a}, with a maximum number of iteration rounds of 100, a batch number of 40, an initialization quantity set to 40, and each group of experiments was run 10 times to count the results. 

Furthermore, the efficiency of the optimization algorithm with and without sizing relations is compared in Table.\ref{tab:comparison}. The cases of optimization failure are also included in the calculation of the average time consumption, and the consumption time of each case is the time consumed until the maximum number of iterations is reached. However, there is still room for improvement in the current scheme. The optimization pass rate still cannot reach 100\%. How to further complement each other between LLM and the optimization algorithm to improve efficiency will be the key to our next research. 
\begin{table}[t]
    \renewcommand{\arraystretch}{1.2}
    \centering
    \caption{Optimization Efficiency and Pass Rate Comparison}
    \resizebox{\linewidth}{!}{
    \begin{tabular}{c >{\columncolor{green!10}}c  >{\columncolor{green!10}}c >{\columncolor{red!10}}c >{\columncolor{red!10}}c} 
        \toprule
         &\multicolumn{2}{c}{\cellcolor{green!10}\textbf{w/ sizing Relationship}}&\multicolumn{2}{c}{\cellcolor{red!10}\textbf{w/o sizing Relationship}}\\ 
         &Time(s)$\downarrow$&Pass Rate$\uparrow$&Time(s)$\downarrow$&Pass Rate$\uparrow$\\ \hline
         \makecell{\textbf{BGR}\cite{ueno2009300}\\@SMIC180nm}&\textbf{3201.3}&\textbf{8/10}&7442&0/10\\
         \makecell{\textbf{OP}\cite{ferreira2007ultra}\\@TSMC65nm}&\textbf{275.17}&\textbf{10/10}&7323&0/10\\
         \makecell{\textbf{LDO}\cite{leung2003capacitor}\\@TSMC65nm}&\textbf{1027.8}&\textbf{10/10}&5081&5/10\\
         \bottomrule
    \end{tabular}
    }
    \label{tab:comparison}
\end{table}
From Table.\ref{tab:comparison}, it can be seen that after adding the sizing relationship, the time consumption of the algorithm decreased by $\OELow \sim \OEHigh \times$ and successfully optimized the circuits that conventional optimization algorithms failed in limited iterations. This experiment further verifies the effectiveness of pruning the search space into the optimization algorithm by bringing the sizing relationships. However, as can be seen from the experiment, although this method has improved the optimization efficiency to a large extent, it still cannot guarantee a 100\% optimization pass rate, thereby increasing the average consumption time.

Overall, through the experiments in this section, we can observe that the extraction of sizing relationships from papers through the multi-agent system proposed in this work not only has strong stability but also significantly improves the efficiency and accuracy of the optimization algorithm by pruning the search space. 

\section{Conclusion}
In this work, we propose an LLM-based multi-agent framework for extracting the analog circuits sizing relationships from academic papers, thereby pruning the search space and ultimately improving the efficiency of the optimization algorithm. We conducted experiments on three types of different circuits and obtained that this method can increase the optimization efficiency of the original optimization algorithm by $\OELow \sim \OEHigh \times$ and successfully optimized the circuits that conventional optimization algorithms failed in limited iterations. 

Under the current background of the shortage of analog circuit data, this method provides a new solution for distilling knowledge from analog circuit academic papers, thereby accelerating the construction process of high-quality data of analog circuits and providing a new scheme for the efficient collection of analog circuit data. 

\bibliographystyle{IEEEtran}
\bibliography{ref.bib}
\end{document}